%% file: 13jmrlarxiv.tex
\documentclass[twoside,11pt]{article}

\usepackage{natbib}
\usepackage{fullpage}

\usepackage{amsmath}
\usepackage{amsfonts}
\usepackage{amssymb}
\usepackage{graphicx}
\usepackage{listings}
\usepackage{colortbl}
\usepackage{enumitem}
\usepackage[font={footnotesize}]{caption}
\usepackage{subcaption}
\usepackage{tabularx}

\usepackage[framemethod=tikz,xcolor=true]{mdframed}
\usetikzlibrary{shadows}

\mdfdefinestyle{exampledefault}{%
    nobreak=true,leftmargin=0.2cm,%
    backgroundcolor=yellow!10,%
    frametitlefont=\scriptsize,%
    roundcorner=5pt,%
    tikzsetting={draw=black, line width=1.0pt},%
}

\newmdenv[style=exampledefault]{Algoritmo}%

\definecolor{darkred}{RGB}{200,0,0}
\definecolor{darkgreen}{RGB}{0,127,0}
\definecolor{purple}{RGB}{127,0,127}
\definecolor{shadow}{RGB}{150,150,150}

\input{macros}

\graphicspath{{figs/}}

\def\Item$#1$[#2]{\item $\displaystyle#1$
   \hfill #2}

\begin{document}

\title{BayesOpt: A Bayesian Optimization Library for Nonlinear Optimization, Experimental Design and Bandits}

\author{Ruben Martinez-Cantin \\
       rmcantin@unizar.es}


\maketitle

\begin{abstract}
BayesOpt is a library with state-of-the-art Bayesian optimization methods to solve nonlinear optimization, stochastic bandits or sequential experimental design problems. Bayesian optimization is sample efficient by building a posterior distribution to capture the evidence and prior knowledge for the target function. Built in standard C++, the library is extremely efficient while being portable and flexible. It includes a common interface for C, C++, Python, Matlab and Octave. 
\end{abstract}


\section{Introduction}

\begin{figure}
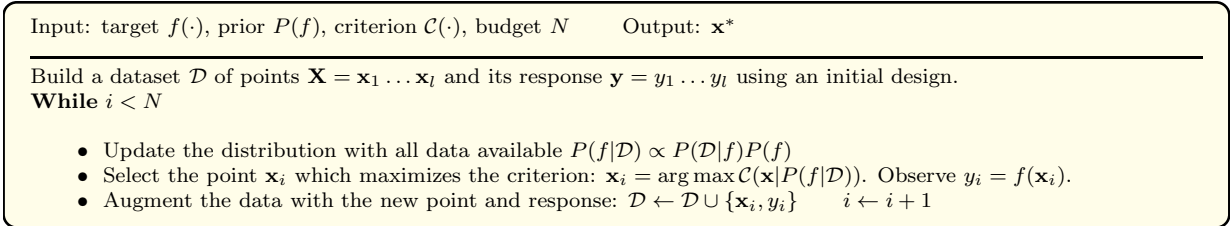

  \centering
  \begin{Algoritmo}[frametitle={Input: target $f(\cdot)$, prior $P(f)$, criterion $\mathcal{C}(\cdot)$, budget $N$ \quad \quad Output: $\x^*$}]
{\scriptsize
\hrulefill\par
Build a dataset $\mathcal{D}$ of points $\X = \x_1 \ldots \x_l$ and its response $\y = y_1 \ldots y_l$ using an initial design.\\
  \textbf{While} $i < N$
  \begin{itemize}[noitemsep]
  \item Update the distribution with all data available $P(f|\mathcal{D}) \propto P(\mathcal{D}|f) P(f)$
  \item Select the point $\x_i$ which maximizes the criterion: $\x_i = \arg \max \mathcal{C}(\x|P(f|\mathcal{D}))$. Observe $y_i = f(\x_i)$.
  \item Augment the data with the new point and response: $\mathcal{D} \leftarrow \mathcal{D} \cup \{\x_i,y_i\} \quad \quad i \leftarrow i+1$
  \end{itemize}}
  \end{Algoritmo}
  \caption{General algorithm for Bayesian optimization}
  \label{fig:algobasic}
\end{figure}

Bayesian optimization \citep{Mockus1989,Brochu:2010c} is a special case of nonlinear optimization where the algorithm \emph{decides} which point to explore next based on the analysis of a distribution over functions $P(f)$, for example a Gaussian process or other \emph{surrogate model}. The decision is taken based on a certain criterion $\mathcal{C}(\cdot)$ called \emph{acquisition function}.  Bayesian optimization has the advantage of having a \emph{memory} of all the observations, encoded in the posterior of the surrogate model $P(f|\mathcal{D})$ (see Figure ~\ref{fig:algobasic}). Usually, this posterior distribution is sequentially updated using a nonparametric model. In this setup, each observation improves the knowledge of the function in all the input space, thanks to the spatial correlation (kernel) of the model. Consequently, it requires a lower number of iterations compared to other nonlinear optimization algorithms. However, updating the posterior distribution and maximizing the acquisition function increases the cost per sample. Thus, Bayesian optimization is normally used to optimize \emph{expensive} target functions $f(\cdot)$, which can be multimodal or without closed-form. The quality of the prior and posterior information about the surrogate model is of paramount importance for Bayesian optimization, because it can reduce considerably the number of evaluations to achieve the same performance. 

\section{BayesOpt library}

BayesOpt uses a surrogate model of the form: $f(\x) = \phi(\x)^T \w + \epsilon(\x)$, where we have $\epsilon(\x) \sim \NP \left( 0, \sigma^2_s (\K(\theta) + \sigma^2_n \I) \right)$.
Here, $\NP()$ means a nonparametric process, for example, a Gaussian, Student-t or Mixture of Gaussians process. This model can be considered as a linear regression model $\phi(\x)^T\w$ with heteroscedastic perturbation $\epsilon(\x)$, as a nonparametric process with nonzero mean function or as a semiparametric model. The library allows to define hyperpriors on $\w$, $\sigma_s^2$ and $\theta$. The marginal posterior $P(f|\mathcal{D})$ can be computed in closed form, except for the kernel parameters $\theta$. Thus, BayesOpt allows to use different posteriors based on empirical Bayes \citep{Santner03} or MCMC \citep{Snoek2012}.

\subsection{Implementation}

\emph{Efficiency} has been one of the main objectives during development. For empirical Bayes (ML or MAP of $\theta$), we found that a combination of global and local derivative free methods such as DIRECT \citep{Jones:1993} and BOBYQA \citep{Powell2009} marginally outperforms in CPU time to gradient based method for optimizing $\theta$ by avoiding the overhead of computing the marginal likelihood derivative. 
Also, updating $\theta$ every iteration might be unnecessary or even counterproductive \citep{Bull2011}.

One of the most critical components, in terms of computational cost, is the computation of the inverse of the kernel matrix $\K(\cdot)^{-1}$. We compared different numerical solutions and we found that the \emph{Cholesky decomposition} method outperforms any other method in terms of performance and numerical stability. Furthermore, we can exploit the structure of the Bayesian optimization algorithm in two ways. First, \emph{points arrive sequentially}. Thus,  we can do incremental computations of matrices and vectors, except when the kernel parameters $\theta$ are updated. For example, at each iteration, we know that only $n$ new elements will appear in the correlation matrix, i.e.: the correlation of the new point with each of the existing points. The rest of the matrix remains invariant. Thus, instead of computing the whole \emph{Cholesky} decomposition, being $\mathcal{O}(n^3)$ we just add the new row of elements to the triangular matrix, which is $\mathcal{O}(n^2)$. Second, finding the optimal decision at each iteration $\x_i$ requires \emph{multiple queries of the acquisition function from the same posterior} $\mathcal{C}(\x|P(f|\mathcal{D}))$ (see Figure \ref{fig:algobasic}). Also, many terms of the criterion function are independent of the query point $\x$ and can be precomputed. This behavior is not standard in nonparametric models, and to our knowledge, this is the first software for Gaussian processes/Bayesian optimization that exploits the idea of precomputing all terms independent of $\x$.

A comparison of CPU time (single thread) vs accuracy with respect to other open source libraries is represented in Table \ref{fig:time} with respect to two different configurations of BayesOpt. SMAC \citep{HutHooLey11-smac}, HyperOpt \citep{Bergstra2011} and Spearmint \citep{Snoek2012} used the HPOlib \citep{EggFeuBerSnoHooHutLey13} timing system (based on \texttt{runsolver}). DiceOptim \citep{Roustant2012} used R timing system (\texttt{proc.time}). For BayesOpt, standard \texttt{ctime} was used.



\begin{table}\centering
{\scriptsize
\renewcommand{\tabcolsep}{3pt}
\begin{tabular}{|l|c|c|r|c|c|r|}
\hline
          & \multicolumn{3}{c|}{Branin (2D)} & \multicolumn{3}{c|}{Camelback (2D)} \\ \hline
          & Gap 50 samp.   & Gap 200 samp.           & Time 200 s.         & Gap 50 samp.            & Gap 100 samp.           & Time 100 s.          \\ \hline
SMAC      & 0.19444 (0.195) & 0.06780 (0.059) & 147.3 (1.3)                    & 0.08534 (0.103) & 0.03772 (0.034) & 70.5 (0.9)  \\ \hline
HyperOpt  & 0.69499 (0.414) & 0.07507 (0.059) & 23.5	(0.2)                   & 0.10941 (0.050) & 0.03383 (0.025) & 8.0 (0.09)   \\ \hline
Spearmint & 1.48953 (1.468) & \textbf{0.00000} (0.000) & 7530.1 (30.4)          & 0.00005 (0.000) & 0.00004 (0.000) & 1674.0 (8.0) \\ \hline
DiceOptim & \textbf{0.00004} (0.000) & 0.00003 (0.000) & 624.3 (35.3) & 0.80861 (0.417) & 0.35811 (0.350) & 215.2 (10.5) \\ \hline
BayesOpt1 & 1.16844 (1.745) & \textbf{0.00000} (0.000) & \textbf{8.6} (0.07)    & 0.00852 (0.021) & \textbf{0.00000} (0.000) & \textbf{2.2} (0.2)  \\ \hline
BayesOpt2 & 0.04742 (0.116) & \textbf{0.00000} (0.000) & 1802.7 (78.3) & \textbf{0.00000} (0.000) & \textbf{0.00000} (0.000) & 147.8 (1.3)  \\ \hline \hline
          & \multicolumn{3}{c|}{Hartmann (6D)} & \multicolumn{3}{c|}{Configuration - $\theta$ learning} \\ \hline
          & Gap 50 samp.          & Gap 200 samp.         & Time 200 s. & \multicolumn{3}{c|}{} \\ \hline
SMAC      & 1.23130 (0.645) & 0.31628 (0.249) & 155.9 (1.3) & \multicolumn{3}{l|}{Default HPOlib} \\ \hline
HyperOpt  & 1.21979 (0.496) & 0.39065 (0.208) & \textbf{33.3} (0.3)& \multicolumn{3}{l|}{Default HPOlib} \\ \hline
Spearmint & 2.13990 (0.659) & 0.59980 (0.866) & 8244.5 (105.8) & \multicolumn{3}{l|}{Def. HPOlib, MCMC (10 particles, 100 burn-in)} \\ \hline
DiceOptim & \textbf{0.06008} (0.063) & 0.06004 (0.063) & 1266.6 (316.4) & \multicolumn{3}{l|}{ML, Genoud 50 pop., 20 gen., 5 wait, 5 burn-in} \\ \hline
BayesOpt1 & 0.06476 (0.047) & \textbf{0.02385} (0.048) & 39.0 (0.04) & \multicolumn{3}{l|}{MAP, DIRECT+BOBYQA every 20 iterations.} \\ \hline
BayesOpt2 & 1.05608 (0.831) & 0.04769 (0.058) & 4093.3 (55.7) & \multicolumn{3}{l|}{MCMC (10 particles, 100 burn-in)} \\ \hline
\end{tabular}}
\caption{Mean (and standard deviation) optimization gap and time (in seconds) for 10 runs for different number of samples (including initial design) to illustrate the convergence of each method. DiceOptim and BayesOpt1 used 5, 5 and 10 points for the initial design, while Spearmint and BayesOpt2 used only 2 points.}
\label{fig:time}
\end{table}

Another main objective has been \emph{flexibility}. The user can easily select among different algorithms, hyperpriors, kernels or mean functions. Currently, the library supports continuous, discrete and categorical optimization. We also provide a method for optimization in high-dimensional spaces \citep{ZiyuWang2013}. The initial set of points (\emph{initial design}, see Figure \ref{fig:algobasic}) can be selected using well known methods such as latin hypercube sampling or Sobol sequences. BayesOpt relies on a factory-like design for many of the components of the optimization process. This way, the components can be selected and combined at runtime while maintaining a simple structure. This has two advantages. First, it is very easy to create new components. For example, a new kernel can be defined by inheriting the abstract kernel or one of the existing kernels. Then, the new kernel is automatically integrated in the library. Second, inspired by the \emph{GPML} toolbox by \cite{Rasmussen2010}, we can easily combine different components, like a linear combination of kernels or multiple criteria. This can be used to optimize a function considering an additional cost for each sample, for example a moving sensor \citep{Marchant2012}. BayesOpt also implements \emph{metacriteria algorithms}, like the bandit algorithm GP-Hedge by \cite{Hoffman2011} that can be used to automatically select the most suitable criteria during the optimization. Examples of these combinations can be found in Section \ref{sec:params}. 

The third objective is \emph{correctness}. For example, the library is thread and exception safe, allowing parallelized calls. Numerically delicate parts, such as the GP-Hedge algorithm, had been implemented with variation of the actual algorithm to avoid over- or underflow issues. The library internally uses NLOPT by \cite{Johnson} for the inner optimization loops (optimize criteria, learn kernel parameters, etc.).

The library and the online documentation can be found at:
 
\centerline{\url{https://bitbucket.org/rmcantin/bayesopt/}}

\subsection{Compatibility}
BayesOpt has been designed to be highly compatible in many platforms and setups. It has been tested and compiled in different operating systems (Linux, Mac OS, Windows), with different compilers (Visual Studio, GCC, Clang, MinGW). The core of the library is written in C++, however, it provides interfaces for C, Python and Matlab/Octave.

\subsection{Using the library}
There is a common API implemented for several languages and programming paradigms. Before running the optimization we need to follow two simple steps:

\subsubsection{Target function definition}
Defining the function that we want to optimize can be achieved in two ways. We can directly send the function (or a pointer) to the optimizer based on a \emph{function template}. For example, in C/C++:
\begin{lstlisting}[language=C]
  double my_function (unsigned int n_query, const double *query, 
                      double *gradient, void *func_data);
\end{lstlisting}
The gradient has been included for future compatibility. Python, Matlab and Octave interfaces define a similar template function.

For a more object oriented approach, we can inherit the abstract module and define the virtual methods. Using this approach, we can also include nonlinear constraints in the \emph{checkReachability} method. This is available for C++ and Python. For example, in C++:
\begin{lstlisting}[language=C++]
class MyOptimization: public bayesopt::ContinuousModel  {
 public:
  MyOptimization(size_t dim, bopt_params param): ContinousModel(dim,param) {}
  double evaluateSample(const boost::numeric::ublas::vector<double> &query) 
  {  // My function here   };
  bool checkReachability(const boost::numeric::ublas::vector<double> &query)
  {  // My constraints here   };
};
\end{lstlisting}

\subsubsection{BayesOpt parameters}
\label{sec:params}
The parameters are defined in the \texttt{bopt\_params} struct --or a dictionary in Python--. The details of each parameter can be found in the included documentation. The user can define expressions to combine different functions (kernels, criteria, etc.). All the parameters have a default value, so it is not necessary to define all of them. For example, in Matlab:

\begin{lstlisting}[language=Matlab]
par.surr_name = 'sStudentTProcessNIG';        % Surrogate model and hyperpriors
% We combine Expected Improvement, Lower Confidence Bound and Thompson sampling
par.crit_name = 'cHedge(cEI,cLCB,cThompsonSampling)';
par.kernel_name = 'kSum(kMaternISO3,kRQISO)';                  % Sum of kernels
par.kernel_hp_mean = [1, 1]; par.kernel_hp_std = [5, 5]; % Hyperprior on kernel
par.l_type = 'L_MCMC';              % Method for learning the kernel parameters
par.sc_type = 'SC_MAP';     % Score function for learning the kernel parameters
par.n_iterations = 200;                       % Number of iterations <=> Budget
par.epsilon = 0.1;          % Add an epsilon-greedy step for better exploration
\end{lstlisting}



\bibliography{../bib/optimizationlong}

\end{document}

%% file: macros.tex
\newcommand{\x}{\mathbf{x}}
\newcommand{\y}{\mathbf{y}}

\newcommand{\I}{\mathbf{I}}
\newcommand{\K}{\mathbf{K}}

\newcommand{\X}{\mathbf{X}}

\newcommand{\w}{\mathbf{w}}

\newcommand{\NP}{\mathcal{NP}}